# DeepER-Med: Advancing Deep Evidence-Based Research in Medicine Through Agentic AI

Zhizheng Wang[1], Chih-Hsuan Wei[1], Joey Chan[1,2], Robert Leaman[1], Chi-Ping Day[3], Chuan Wu[4], Mark A Knepper[5], Antolin Serrano Farias[6], Jordina Rincon-Torroella[6], Hasan Slika[2], Betty Tyler[2], Ryan Huu-Tuan Nguyen[7,8], Asmita Indurkar[9], Mélanie Hébert[9], Shubo Tian[1], Lauren He[1], Noor Naffakh[8], Aseem Aseem[8], Nicholas Wan[10], Emily Y Chew[9], Tiarnan D L Keenan[9], Zhiyong Lu[1*]

[1] Division of Intramural Research (DIR), National Library of Medicine (NLM), National Institutes of Health (NIH), Bethesda, MD 20892, USA

[2] Hunterian Neurosurgical Laboratory, Johns Hopkins School of Medicine, Baltimore, MD 21231, USA

[3] Cancer Data Science Laboratory, Center for Cancer Research, National Cancer Institute (NCI), National Institutes of Health (NIH), Bethesda, MD 20892, USA

[4] Experimental Immunology Branch, National Cancer Institute (NCI), National Institutes of Health (NIH), Bethesda, MD 20892, USA

[5] Epithelial Systems Biology Laboratory, Systems Biology Center, National Heart, Lung, and Blood Institute (NHLBI), National Institutes of Health (NIH), Bethesda, MD 20892, USA

[6] Brain Tumor Genetics Lab, Department of Neurosurgery, Johns Hopkins Hospital, Baltimore, MD 21287, USA

[7] Division of Hematology/Oncology, University of Illinois Chicago, Chicago, IL 60612, USA

[8] University of Illinois Cancer Center, Chicago, IL 60612, USA

[9] Division of Epidemiology and Clinical Applications, National Eye Institute (NEI), National Institutes of Health (NIH), Bethesda, MD 20892, USA

[10] University of Michigan Medical School, Ann Arbor, MI 48109, USA

\* Correspondence: zhiyong.lu@nih.gov

## Abstract

Trustworthiness and transparency are essential for the clinical adoption of artificial intelligence (AI) in healthcare and biomedical research. Recent deep research systems aim to accelerate evidence-grounded scientific discovery by integrating AI agents with multi-hop information retrieval, reasoning, and synthesis. However, most existing systems lack explicit and inspectable criteria for evidence appraisal, creating a risk of compounding errors and making it difficult for researchers and clinicians to assess the reliability of their outputs. In parallel, current benchmarking approaches rarely evaluate performance on complex, real-world medical questions. Here, we introduce **DeepER-Med**, a **Deep Evidence-based Research** framework for **Medicine** with an agentic AI system. DeepER-Med frames deep medical research as an explicit and inspectable workflow of evidence-based generation (EBG), consisting of three modules: research planning, agentic collaboration, and evidence synthesis. To support realistic evaluation, we also present **DeepER-MedQA**, an evidence-grounded dataset comprising 100 expert-level research questions derived from authentic medical research scenarios and curated by a multidisciplinary panel of 11 biomedical experts. Expert manual evaluation demonstrates that DeepER-Med consistently outperforms widely used production-grade platforms across multiple criteria, including the generation of novel scientific insights. Beyond manual assessment, we further evaluate DeepER-Med across distinct stages of EBG using quantitative metrics, including semantic similarity and information entropy, capturing both system performance and the relevance of retrieved evidence. We further demonstrate the practical utility of DeepER-Med through eight real-world clinical cases. Human clinician assessment indicates that DeepER-Med's conclusions align with clinical recommendations in seven cases, highlighting its potential for medical research and decision support.

# INTRODUCTION

Different from regular question-answering (QA) system that retrieve information once and generate a single-pass answer, deep research is a class of agentic artificial intelligence (AI) capabilities and models that conduct multi-step information retrieval across the web and databases, analyze findings, and generate citation-based reports with transparent reasoning[1]. Recent advances in agentic AI further support deep research by enabling systems to perform contextually relevant actions in response to complex research questions[2,3]. Such agentic systems are increasingly being suggested for healthcare research by coordinating large language models (LLMs) and specialized AI agents for literature review and analysis[4,5,6], or knowledge synthesis and reasoning[7,8,9]. Production-grade platforms including OpenAI Deep Research[10], Google AI Mode (Deep Search)[1], and Open Evidence[11] are increasingly adopted by the scientific community and healthcare professionals to seek evidence-based answers, given their ability to perform in-depth research by associating supporting evidence alongside model generated responses.

However, the real-world performance of these existing systems in comprehensive literature exploration and expert-aligned evidence synthesis remains insufficiently examined, which limits confidence in their practical use in medicine[1,12]. Most deep research systems are evaluated using simplified questions drawn from open-access databases or extracted from scientific studies[13,14]. Such benchmarks primarily emphasize multiple-choice answer accuracy without the verification of supporting references, providing limited insight into whether system performance reflects genuine evidence interpretation or instead derives largely from the underlying capabilities of backend LLMs. Critically, these evaluations rarely capture the demands of frontline medical research in real-world scenarios, where trustworthiness, interpretability, and evidence reliability are essential.

In parallel, existing deep research systems typically follow an agent-loop setting that iterates through query understanding, tool calls, and context updates[15,16]. While this loop enables iterative retrieval, it risks amplifying uncertainty and compounding errors across iterations, as early misinterpretations or incomplete retrieval can become locked into subsequent searches and summaries. Moreover, the intermediate processes of evidence selection, aggregation, and interpretation are frequently opaque to researchers, making it difficult to determine whether the final conclusions reflect robust evidence synthesis (e.g., transparent inclusion criteria and quality appraisal) or instead a superficial narrative assembled from repeatedly compressed paraphrases of underlying sources.

To address these limitations, we propose a deep evidence-based research paradigm that unlocks evidence-based generation (EBG) for medical scientific discovery, which is inspired by evidence-based medicine in the healthcare domain[17,18,19]. Specifically, we develop **DeepER-Med**, an EBG–driven agentic AI framework that operates through three integrated modules: research planning for intent identification, agentic collaboration for evidence retrieval, appraisal, and interpretation, and evidence synthesis. In this way, it frames medical research as an "evidence distillation" process that begins with thoroughly retrieving evidence for all identified research intents and is followed by criteria-driven evidence appraisal and refinement, thereby reducing the risk of compounding errors. Moreover, user inputs and constraints explicitly guide evidence synthesis and answer generation, ensuring that the resulting conclusions remain tightly aligned with the scope and priorities of the research question. Critically, DeepER-Med retrieves references directly from source databases (e.g. ClinicalTrials.gov or PubMed) instead of generating them via LLMs, ensuring that supporting references are grounded and verifiable. In contrast, hallucinated citations remain a limitation in other systems and pose a significant risk to the scientific ecosystem (cite the nature article).

---

[1] https://support.google.com/websearch/answer/16420827?hl=en

To enable rigorous evaluation of DeepER-Med and other similar systems, we further introduce **DeepER-MedQA**, a benchmark of 100 expert-level questions designed to capture the complexity of real-world medical research. DeepER-MedQA was curated by a multidisciplinary panel of 11 biomedical domain experts, including both senior investigators and early-career researchers, who contributed research questions and participated in the evaluation of AI-generated outputs. Beyond answer accuracy, expert assessment measures the comprehensiveness of evidence as well as the quality and coherence of evidence interpretation, enabling a multidimensional evaluation of system performance. Across these criteria, DeepER-Med shows consistently stronger performance than three production-grade, cutting-edge systems: OpenAI Deep Research, OpenEvidence, and Google AI Mode (Deep Search). These systems were selected based on their state-of-the-art performance, which substantially surpasses that of LLM-only or RAG-based approaches[20,21,22], as well as their growing use for evidence-grounded scientific and clinical inquiry.

To complement resource-intensive expert evaluation, we further conduct a large-scale assessment of DeepER-Med across distinct stages of EBG-style deep research in medicine using five public biomedical datasets spanning contextual QA, open-ended QA, attribution QA, and the hypothesis verification task. These datasets allow isolation of performance in intent resolution, literature retrieval, evidence interpretation, and knowledge summarization. System performance is quantified using both semantic similarity and information entropy, enabling joint assessment of prediction correctness and the diversity and alignment of retrieved evidence. Together, these analyses indicate that DeepER-Med achieves high task-specific accuracy while maintaining strong evidence relevance and controlled expansion of the evidentiary landscape.

Finally, to assess its practical utility in clinical contexts, we evaluate DeepER-Med on real-world oncology cases requiring integration of heterogeneous patient data, up-to-date literature retrieval, and evidence-grounded analysis of treatment-related questions. Clinician review indicates that in most cases the output of DeepER-Med is consistent with expert recommendations and supported by traceable evidence.

# RESULTS

## Overview of the DeepER-Med framework

We developed DeepER-Med to support evidence-based generation paradigm in deep medical research by linking evidence retrieval, appraisal, interpretation, and synthesis into a structured knowledge summarization workflow for medical scientific discovery (**Fig. 1a,b**). The system operates through three integrated modules: research planning, agentic collaboration, and evidence synthesis. These three modules involve five specific stages: research intent identification, evidence retrieval, appraisal, interpretation, and synthesis.

Given a question, the framework first generates a hierarchical research plan that decomposes the query into a sequence of sub-questions progressing from foundational concepts to higher-order relationships. For example, when asked the question "What are the earliest imaging features of macular telangiectasia type 2?", DeepER-Med first identifies relevant imaging modalities, then examines modality-specific findings before inferring early disease features. This stepwise decomposition provides an interpretable representation of research intent, while simultaneously broadening the scope of the subsequent evidence retrieval.

The framework next thoroughly retrieves relevant evidence for each sub-question by integrating multiple biomedical information sources, including medical knowledge graphs, literature databases, and domain-specific resources (**Fig. 1b**). A coordinated agentic collaborative workflow determines whether a sub-question requires relational reasoning—triggering knowledge-graph expansion—or direct literature retrieval. By integrating heterogeneous evidence sources within such a hierarchical agentic collaborative workflow, DeepER-Med enables the retrieval and incorporation of real-world evidence in clinical trial data in addition to the biomedical literature. Retrieved evidence is then screened according to predefined criteria emphasizing contextual relevance, evidentiary strength, and methodological quality, which contributes to the transparency and inspection of evidence appraisal and inclusion.

Finally, the framework synthesizes the information retrieved into a coherent response to the original question. Candidate evidence is further filtered against the user's specific research constraints and objectives and integrated into a concise response accompanied by a structured analytical report grounded in traceable references. Notably, references are retrieved via direct queries to the source database API rather than generated by LLMs, ensuring that the reference list remains verifiable and free from hallucination. Detailed implementations of DeepER-Med are described in the **Methods** section.

## DeepER-MedQA benchmark design

To evaluate evidence-aware deep research systems in realistic biomedical settings, we developed DeepER-MedQA, an expert-curated benchmark of complex medical research questions (**Fig. 1c**). DeepER-MedQA was constructed through a three-stage pipeline comprising draft question generation, human expert annotation, and reference-answer curation (**Methods**).

Domain experts contributed 100 complex questions spanning 11 disease-related research areas, including melanoma, glioblastoma, and age-related macular degeneration. (**Fig. 2a, left**). These questions were designed to reflect real-world biomedical research inquiry ranging from mechanistic investigations to translational and clinical medical decision-making tasks. Candidate answers to each question were generated by DeepER-Med and three widely used production-grade systems: OpenAI Deep Research, Open Evidence; and Google AI Mode (Deep Search). Responses of these systems were standardized to include a concise answer, analytical report, and supporting references to enable consistent evaluation.

Expert reviewers assessed blinded responses along five dimensions: factual accuracy, analytical coherence, reference quality, novelty of insights, and comprehensiveness. High-quality responses were identified and combined as candidate reference answers. These selected answers were further verified through an iterative human-AI review pipeline to ensure evidence traceability and factual correctness (**Methods**).

To characterize their scientific focus, questions were categorized into three domains (**Methods**). Fifty-four percent addressed disease-related biological mechanisms at the molecular or cellular level (i.e., basic research). Twenty-two percent examined translational problems linking biological mechanisms to diagnostic, prognostic, or therapeutic applications (i.e., translational research). The remaining 24% focused directly on clinical questions (i.e., clinical research) (**Fig. 2a, right**).

Each benchmark entry includes five components: question category, question text, curated reference answer, expert annotations, and supporting references (**Fig. 2b; Extended Table 1**). Supporting references are ordered chronologically to allow users to trace the development of the evidence base underlying each answer.

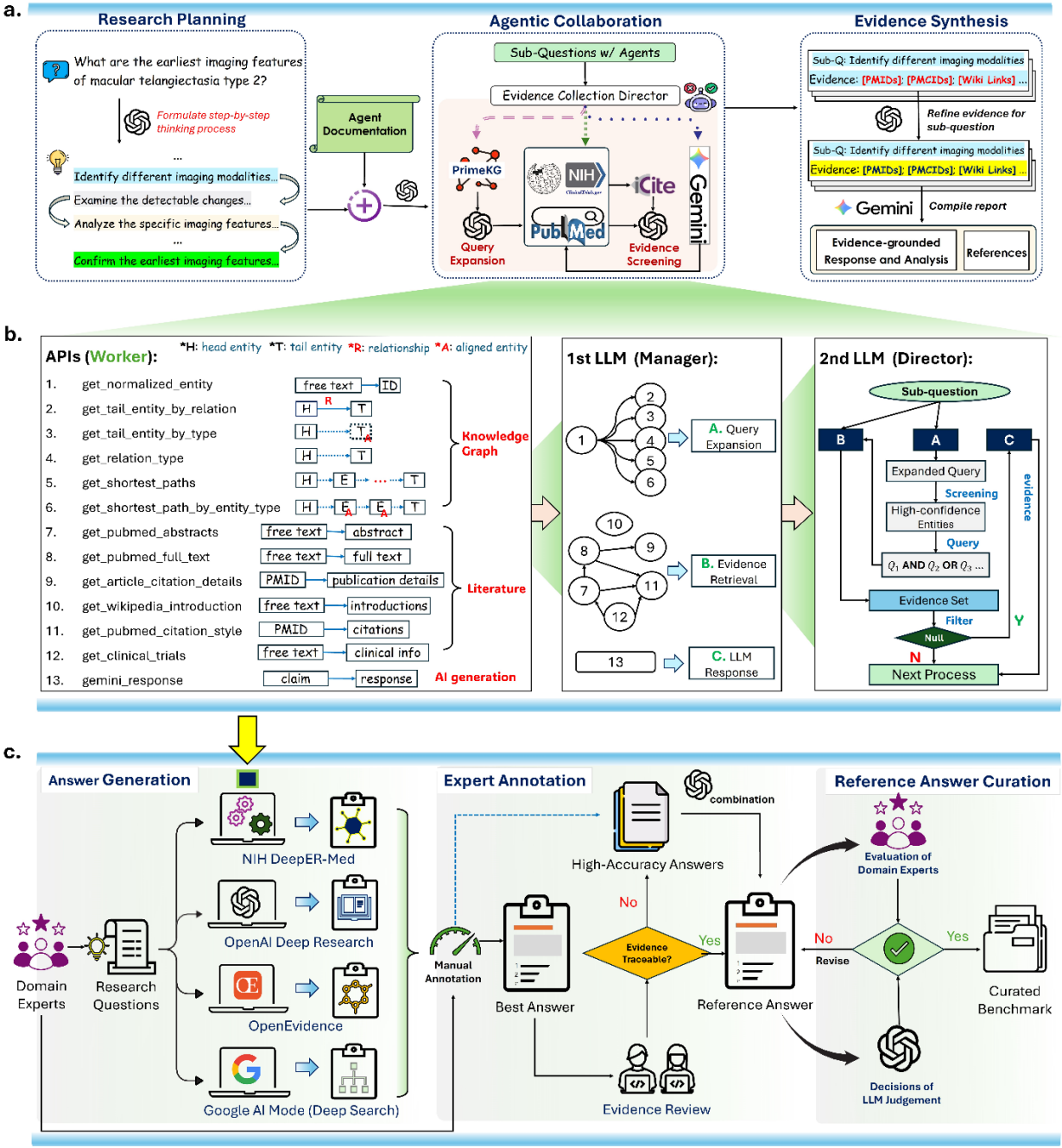

**Figure 1. Overview of the DeepER-Med framework and the DeepER-MedQA curation.** **A.** Workflow of DeepER-Med, consisting of research planning, agentic collaboration, and evidence synthesis. Different LLMs are employed at distinct stages of the workflow, including GPT-based models for planning and evidence interpretation and Gemini-based models for evidence synthesis. **B.** Three-layer agentic collaboration network that retrieves, filters, and interprets evidence by organizing heterogeneous APIs into different functionally specialized agent communities. **C.** Pipeline of answer curation for DeepER-MedQA, consisting of three stages: candidate answer generation, expert annotation, and reference answer curation. Human expert evaluation is integrated with LLM-based judgment (GPT-5.2-pro) to validate reference answers through an iterative human–AI consensus process (**Methods**).

**Benchmarking DeepER-Med against major deep research systems**

To evaluate performance, 11 domain experts assessed responses generated by DeepER-Med and three cutting-edge systems—OpenAI Deep Research, OpenEvidence, and Google AI Mode (Deep Search)—across all questions curated in DeepER-MedQA (**Methods**).

Across all evaluation dimensions, DeepER-Med consistently outperformed the comparator systems. The largest improvements were observed in reference relevance (81 cases versus 59 cases for the strongest baseline) and analytical quality (67 cases versus 47 cases), indicating stronger evidence retrieval and integration (**Fig. 2c**). These improvements were reflected in answer accuracy: DeepER-Med produced highly accurate responses in 77 cases compared with 69 cases for the best-performing comparator. Experts also highlighted the ability of DeepER-Med to generate new insights grounded in evidence. In 27 cases, reviewers preferred DeepER-Med responses because they identified previously unrecognized connections supported by the literature (**Fig. 2c**). Consistent with this observation, DeepER-Med received higher comprehensiveness scores, suggesting that it synthesized a broader set of relevant evidence (**Fig. 2d**).

When experts selected the best response among systems for each question, DeepER-Med was chosen in 60 cases, including 27 sole selections (**Fig. 2e**). The strongest baseline system was selected in 43 cases (19 sole selections). OpenEvidence and Google AI Mode (Deep Search) were selected less frequently and typically shared selections with other systems, indicating weaker standalone performance. Evaluation of reference lists further highlighted DeepER-Med's advantage in evidence retrieval (**Fig. 2f**). DeepER-Med provided the most relevant and well-supported references and retrieved a broader literature base per question. Analysis of publication-year metadata showed that cited studies primarily spanned 2015–2025, indicating both breadth and recency of evidence coverage (**Extended Fig. 1**).

Qualitative review of expert annotations revealed systematic differences in reasoning structure across systems (**Extended Fig. 2**). Comparator systems frequently produced outputs characterized as broad explanations or superficial summaries, suggesting that retrieved information was often reformulated without being integrated into coherent analytical reasoning. Lower-quality responses were also associated with missing citations, weak evidence support, or imprecise language. In some cases, comparator systems also generate fabricated citations for their claims, further undermining the reliability of their conclusions.

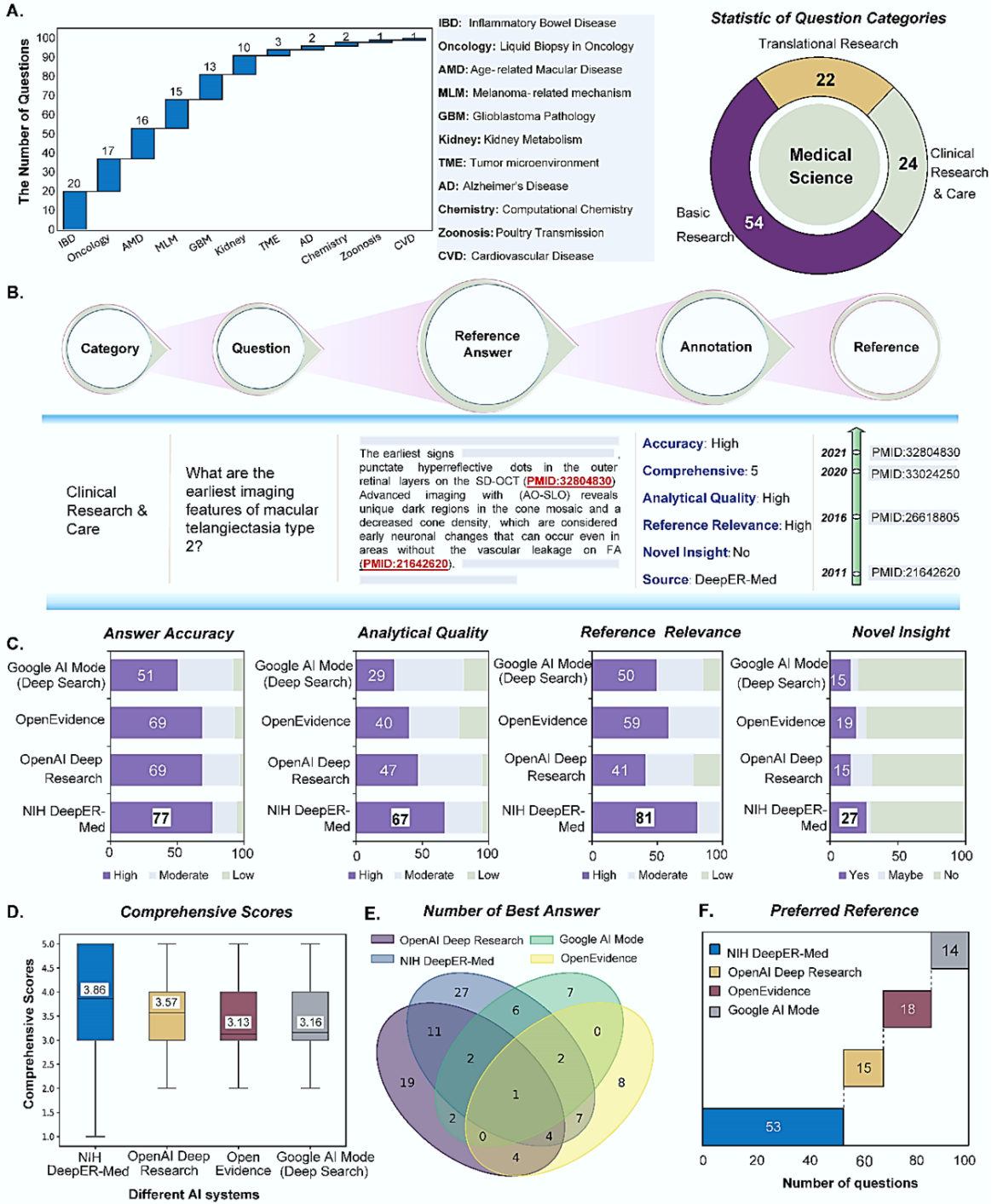

**Figure 2. Overview of DeepER-MedQA and expert evaluation of DeepER-Med. A.** Distribution of questions in DeepER-MedQA by research topics (n = 100). **B.** Example of a curated DeepER-MedQA entry. **C, D.** Results of expert manual evaluation of DeepER-Med and comparator evidence-aware systems (n = 100). Boxplots show the median and interquartile range (IQR); whiskers extend to the most extreme values within 1.5× IQR. **E.** Expert selection frequencies of best-performing responses among the four evidence-aware systems. **F.** Expert preference analysis of reference lists generated by each system.

**Mechanistic analysis of evidence retrieval and synthesis in DeepER-Med**

To examine how the proposed evidence-based generation paradigm contributes to performance, we analyzed DeepER-Med across the major components of its workflow: intent identification, literature retrieval, evidence interpretation, and knowledge synthesis. These four configurations were evaluated using five open biomedical datasets spanning question-answering and hypothesis verification tasks (**Tab. 1**).

**Intent identification**

Accurate identification of research intent is critical because downstream retrieval and synthesis depend on correctly interpreting the question. DeepER-Med addresses this stage through hierarchical sub-question decomposition.

Based on the PubMedQA dataset (**Tab. 1**) that provides "context" where questions are curated from, ablation experiments demonstrated that DeepER-Med achieved **79.2% accuracy** when both contextual information and sub-question decomposition were used (**Fig. 3a**). Removing contextual information produced only a minor decrease (**78.2%**), whereas removing sub-questions reduced accuracy to 74.4%. These results indicate that intent identification via hierarchical sub-question decomposition substantially improves decision accuracy, a benefit attributed to the ability to gather a broader and deeper range of evidence when answering questions.

**Literature retrieval**

The literature retrieval stage emphasizes identifying evidence that is both research relevant and sufficiently comprehensive. Retrieval relevance was evaluated using the BioMaze dataset, which contains open-ended biomedical question-answering tasks across six domains (**Tab. 1**).

We embedded the set of reference publications and the set of literature retrieved by DeepER-Med using MedCPT[23] (**Methods**) and computed semantic similarity between the two sets. Retrieved literature consistently showed strong alignment with the reference publications, with similarity scores exceeding 70% across domains (**Fig. 3b**). Projecting the embedding into a two-dimensional space further revealed that DeepER-Med-retrieved literature occupied a broader semantic region than the reference publications, while remaining centered within the same topic clusters (**Fig. 3c**). This pattern holds across all domains, suggesting that the proposed DeepER-Med expands the evidentiary landscape while maintaining topical alignment.

**Evidence interpretation**

The interpretation stage evaluates whether retrieved evidence is appropriately aligned with evidence curated for system generations. This stage was assessed using attribution question answering and hypothesis verification datasets (**Tab. 1**), in which conclusions derived from explicit source spans.

For attribution QA tasks, we measured evidence diversity using Shannon information entropy[24,25,26]. DeepER-Med exhibited higher entropy than multi-model aggregated evidence and values closer to expert-curated evidence (**Fig. 3d, left**), indicating broader informational diversity. We also computed

Jensen–Shannon divergence[27,28] between DeepER-Med-retrieved evidence and expert-curated evidence (**Fig. 3d, right**). Divergence values were low (0.04–0.05), indicating strong distributional alignment. Together, these findings suggest that DeepER-Med retrieves evidence that is both diverse and closely aligned with expert expectations (**Extended Fig. 3a**). For the hypothesis verification task, we compared ground-truth source studies with those cited in DeepER-Med outputs. Retrieved literature covered **96.5%** of the source studies (**Fig. 3e, left**), indicating very high recall of relevant evidence. Consequently, 91.3% hypotheses were correctly verified as "true", while only 8.7% true hypotheses were verified as "false" (**Fig. 3e, right**).

**Knowledge synthesis**

Finally, we evaluate the accuracy of knowledge synthesis across the BioMaze and PubMedQA datasets (**Tab. 1**). Both datasets include qualitative answers to the questions that can be assessed through semantic matching.

On BioMaze, DeepER-Med outperformed several comparator configurations including chain-of-thought prompting, retrieval-augmented generation, and graph-augmented LLM methods such as G-Retriever[29] and CoK[30]. For binary-choice questions, DeepER-Med improves accuracy by 7% relative to GPT-4o with chain-of-thought prompting. For open-ended QA, DeepER-Med achieved 90.6% accuracy (**Fig. 3f**). Open-ended responses evaluated using GPT-5.2 as judge under predefined prompts showed 90% agreement with human annotations (**Extended Fig. 3b**). On PubMedQA, DeepER-Med achieved 79.2% accuracy, comparable to advanced retrieval-augmented approaches including MedPrompt[31] (82.0%), MedReason[32] (79.4%), and Flan-PaLM[33] (79.0%).

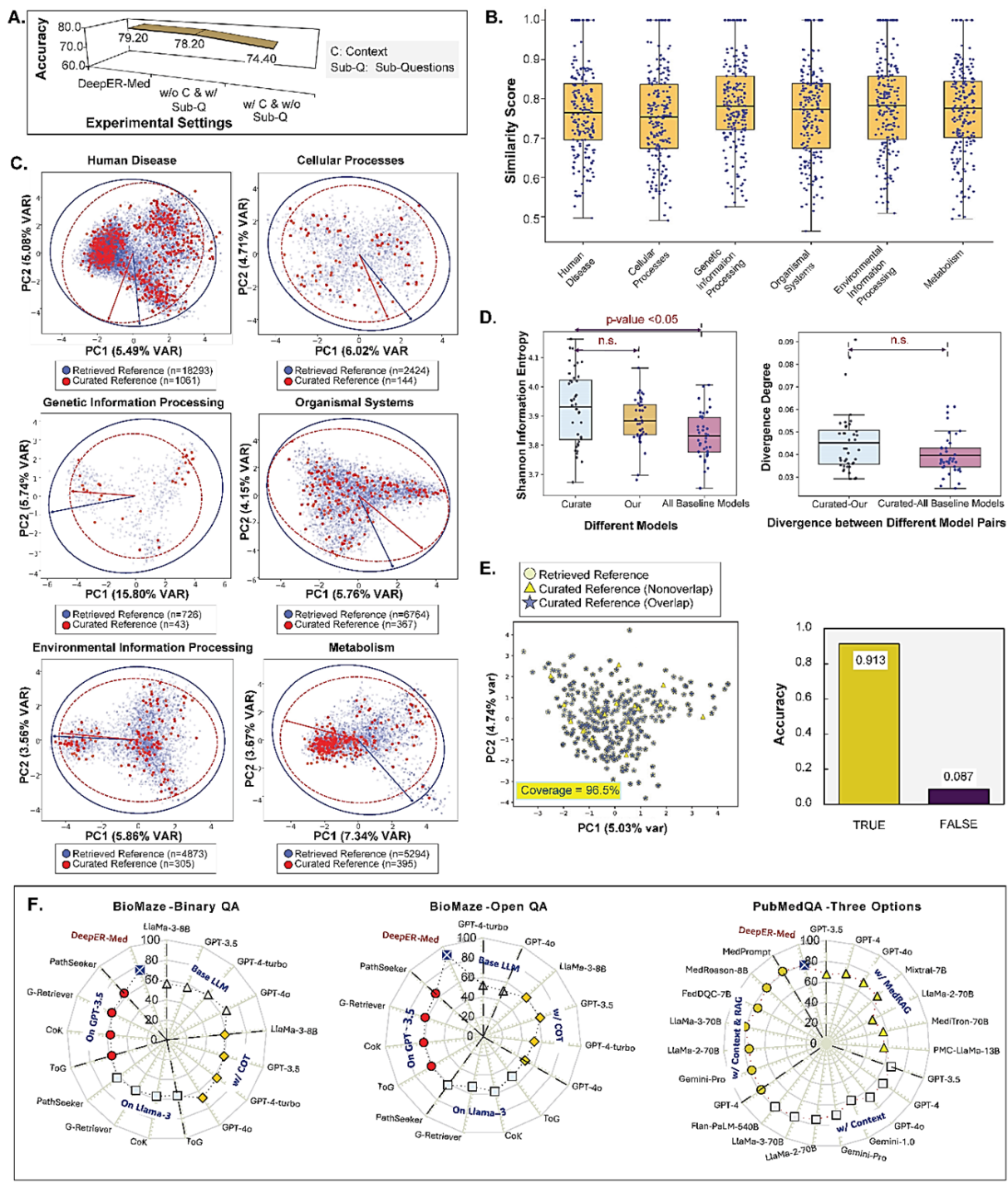

**Figure 3. Automatic evaluation of DeepER-Med on open-access datasets. A.** Ablation experiments on PubMedQA (i.e., with or without sub-question decomposition). **B.** Semantic similarity between reference publications and literature retrieved by DeepER-Med for the open-ended QA task, computed using MedCPT embeddings. **C.** Comparison of the conceptual scope of DeepER-Med-retrieved literature and reference publications for the open-ended QA task, visualized through 2D embedding projections. **D.** Evidence diversity and alignment analysis for the attribution QA task, showing information entropy of expert-curated evidence, DeepER-Med-retrieved evidence, and multi-model aggregated evidence (left), together with Jensen–Shannon

divergence between distributions of system-generated evidence and curated evidence (right). P-value is computed by paired two-tails T Test. "n.s." denotes no significant difference. **E.** Coverage of source studies by literature cited in DeepER-Med-generated answers for the hypothesis verification task (left), reflecting the extent to which system outputs capture ground-truth supporting research, along with accuracy of DeepER-Med in hypothesis verification (right). **F.** Accuracy of DeepER-Med on QA tasks across BioMaze and PubMedQA datasets. Boxplots in **B** and **D** indicate the median and interquartile range (IQR); whiskers extend to the most extreme values within 1.5× IQR.

**Table 1. Datasets used for mechanistic evaluation of DeepER-Med.** Except for BioDSA, which is a subset of the BioDSA-1K dataset, the other four datasets contain raw data without any processing.

| Dataset | Count | Evaluation | Specificity for Evaluation | Example |
|---|---|---|---|---|
| PubMedQA[34] | 500 | Intent identification & Knowledge synthesis | Besides answers with **"yes, no. or maybe"**, it also includes the **"context"** of the question, so as to facilitate comparison with the **"sub-question"**. | **Question:** Can transcranial direct current stimulation be useful in ....? **Answer:** Yes **Context:** ['Disorders of consciousness (DOC) diagnosis relies on the presence or absence of purposeful motor responsiveness, …'] |
| BioMaze-Open QA[35] | 2,523 | Literature retrieval & Knowledge synthesis | Questions and supporting evidence are categorized into **predefined domains**, so as to evaluate the **topic relevance of evidence**. It also provides **ground-truth summaries** based on evidence. | **Question:** In primary AML blasts, What is the effect of the AML1-ETO, …? **Answer:** AML1-ETO, PML-RARalpha, and PLZF-RARalpha fusion proteins significantly overexpress plakoglobin in primary AML blasts. **Biomedical Category:** Human Diseases |
| MedAESQA[36] | 40 | Evidence interpretation | Each answer is supported by **multiple, distinct evidence sources**, so as to assess both **evidence diversity and distributional alignment** in the evidence interpretation stage. | **Question:** Are there ways to prevent sleep apnea or treat it naturally? **Curated Answer:** There are ways to prevent … *[28646811, 33659106, 28659501, 12693795, 36617387]*. Sleep apnea can be …. *[33659106, 12693795, 30204000, 36617387, 28659501]*; …. **Nuggets:** ["Exists (treat naturally[C0027495], sleep apnea [C0037315])", "[may] Prevent (lifestyle changes [C0870811]", … ] |
| BioDSA*[37] | 520 | Evidence interpretation | Each answer is traceable to **exact source spans**, so as to assess **evidence consistency** in the evidence interpretation stage. | **Question:** Amplification of MYC is associated with poor outcome in pancreatic ductal adenocarcinoma? **Answer:** True **Source:** PMID:25855536 |
| BioMaze-Binary QA[35] | 2,622 | Knowledge synthesis | Answers with **"yes or no"**, which can serve as a **complementary evaluation setting** to BioMaze Open QA in the knowledge synthesis stage. | **Question:** In the context of …, does the inactivation of magnesium ion (Mg(2+)) …? **Answer:** Yes **Biomedical Category:** Organismal Systems |

## Case Study in Precision Oncology Tumor Board (POTB) Settings

To illustrate the potential utility of DeepER-Med in real-world clinical contexts, we conducted a case study using discussions from a Precision Oncology Tumor Board (POTB) at the University of Illinois Cancer Center. POTBs convene oncologists, pathologists, geneticists, translational researchers, and clinical pharmacists to review complex molecular and clinical data and develop patient-specific treatment strategies[38].

We analyzed eight clinical cases derived from POTB discussions. Each case included detailed clinical histories, biomarker measurements, genomic alterations, and research questions reflecting real clinical uncertainties. DeepER-Med outputs were reviewed by clinicians involved in POTB activities and evaluated according to two criteria: (i) evidence reliability and (ii) consistency with POTB recommendations (**Methods**).

Across the evaluated eight cases (**Tab. 2**), DeepER-Med's conclusions were judged consistent with previous human-expert recommendations of POTB in **seven cases** (CID:1–CID:6 and CID:8). The remaining case involved a clinically nuanced scenario with evolving evidence and was considered discordant. Evidence reliability was judged fully satisfactory in **five cases** and partially satisfactory in the remaining cases due to incomplete coverage of prior studies. Notably, even among partially reliable cases, DeepER-Med's conclusions aligned with POTB recommendations in two instances, suggesting that it has included key evidence for clinical diagnosis. These findings demonstrate that DeepER-Med can assist clinicians by organizing heterogeneous biomedical evidence and synthesizing relevant literature in complex oncology settings. Rather than replacing multidisciplinary decision-making, such systems may support clinical discussion by identifying relevant studies and structuring evidence-intensive analyses.

**Table 2. Evaluations of DeepER-Med on eight real-world clinical cases.** Summary of clinician evaluations of DeepER-Med-generated responses across eight clinical cases, assessed along criteria of clinical consistency and evidence reliability. The final annotations reflect consensus judgments from three participating clinicians following discussion within their research communities. (NSCLC: non–small cell lung cancer; HG NEC / SCLC: High-grade neuroendocrine carcinoma consistent with small cell lung cancer; mCSPC: metastatic castration-sensitive prostate cancer; TMB: tumor mutational burden.)

| Clinical ID | Clinical Topic | Evidence Reliability | | POTB's Recommendations? | |
|---|---|---|---|---|---|
| | | Complete | Partial | Consistent | Discordant |
| CID: 1 | NSCLC | ✓ | | ○ | |
| CID: 2 | NSCLC | ✓ | | ○ | |
| CID: 3 | HG NEC / SCLC | ✓ | | ○ | |
| CID: 4 | NSCLC | ✓ | | ○ | |
| CID: 5 | Suspected Lung Cancer | | ✓ | ○ | |
| CID: 6 | NSCLC w/ ALK Rearrangement | | ✓ | ○ | |
| CID: 7 | mCSPC w/ NBN Mutation | | ✓ | | ○ |
| CID: 8 | mCSPC w/ TMB-High | ✓ | | ○ | |

# DISCUSSIONS

## Transparency and trustworthiness

This study introduces evidence-based generation (EBG) as a framework for structuring deep medical research with AI systems and evaluates it through the joint development of an evidence-grounded agent (DeepER-Med) and an expert-curated benchmark (DeepER-MedQA). These components shift the focus of AI-assisted biomedical research from task-level answer accuracy of regular LLM-only or RAG-based approaches toward the transparency and reliability of the deep research setting.

Our work elevates evidence transparency to a primary research outcome. Evaluations show that DeepER-Med achieves high accuracy without compromising the traceability of evidence selection (**Fig. 2, Fig 3**). These improvements suggest that multi-stage workflows are more reliable than end-to-end reasoning pipelines. Similarly, the curation of DeepER-MedQA illustrates how human–AI interaction enhances trust without sacrificing scalability (**Fig. 1c**). By using expert criteria and iterative validation, our framework prioritizes human-in-the-loop oversight, treating human expertise and automated evaluation as complementary rather than competing. Under expert supervision, DeepER-MedQA reduces hallucinations while preserving transparency.

## Comparison with AI agents

By focusing on comparisons with existing production systems, our results suggest that current limitations arise less from restricted access to information than from the difficulty of prioritizing, integrating, and interpreting evidence. Although modern retrieval systems can access large volumes of biomedical literature, expert evaluation showed that competing platforms frequently condensed retrieved information into broad explanations or superficial summaries rather than integrating it into coherent analytical reasoning (**Extended Fig. 2**). In contrast, DeepER-Med explicitly models different research intents and evidence selection criteria, enabling more structured reasoning grounded in supporting studies. Moreover, rather than relying on broad web searches (e.g., Google AI mode) that may include unverified medical information, DeepER-Med prioritizes curated, trustworthy medical resources to ensure the reliability and accountability of the evidence it retrieves.

Recent agentic biomedical research platforms such as Biomni[39], DeepEvidence[40], and Alvessa[41] have demonstrated the feasibility of coordinated tool-based reasoning, but their evaluations primarily emphasize black-box task-level accuracy (**Append Table 1**). By pairing an inspectable agentic workflow with a consensus-driven benchmark, our approach repositions the primary objective of AI-assisted discovery from mere result generation to process-level accountability. By decomposing research intent and enforcing explicit inclusion criteria, the system effectively "opens the black box," transforming scientific synthesis from an opaque heuristic into a transparent, auditable sequence of evidence-grounded decisions. This shift is critical for research governance: it allows practitioners not only to evaluate the accuracy of a final claim but to quantitatively verify the information-theoretic rigor and expert alignment of the underlying evidence.

## Expanded evidence depth and breadth

A key component of DeepER-Med is the structured exploration of research intent—operationalized through sub-question decomposition of increasing complexity—which contributes to the evidence depth. Ablation experiments (**Fig. 3a**) indicate that this step contributes substantially to decision accuracy, and in some cases it enabled identification of more recent or relevant evidence than contextual summaries alone. For example, when addressing the question of whether T-cell deficiency affects spatial learning following toluene exposure, DeepER-Med retrieved more recent supporting evidence (PMID:36411380[42], 2022) that refined the interpretation of earlier findings (PMID:19923859[43], 2010). Such cases highlight how structured exploration facilitates a more layered and informed evidentiary grounding, thereby expanding the depth of evidence underlying a research question.

The retrieval analyses also suggest that DeepER-Med expands the evidentiary landscape in a controlled manner (**Fig. 3b,c,d**). Our semantic similarity and information-theoretic analyses show that retrieved literature remains closely aligned with expert-curated evidence while covering a broader portion of the relevant research space. The combination of increased information entropy with low Jensen–Shannon divergence indicates that the system balances breadth and relevance rather than indiscriminately expanding the evidence pool. The implications of this balance extend beyond system performance. A recent report found that while AI-augmented research tools increase scientific productivity, they simultaneously narrow the diversity of topics collectively explored by researchers[44]. By retrieving a broader and more diverse set of studies while maintaining alignment with expert expectations, DeepER-Med partially addresses concerns that AI-assisted discovery concentrates scientific attention on a narrower subset of the literature. This controlled expansion may be particularly important for exploratory biomedical research, where identifying less obvious connections across studies can generate new hypotheses.

Importantly, improvements in upstream reasoning stages translated into accurate knowledge synthesis without relying on complex prompting strategies or ensemble voting. On PubMedQA, DeepER-Med achieved performance comparable to advanced retrieval-augmented systems despite using a simpler inference pipeline (**Fig. 3f**). These findings suggest that modeling research intent and controlling evidence expansion may contribute more to robust biomedical reasoning than increasingly elaborate prompt engineering. Ablation experiments (**Extended. Fig. 3c**) further suggest that query expansion guided by structured knowledge graphs contributes substantially to system performance, reinforcing the importance of modeling research intent and domain constraints through structured biomedical knowledge.

## Expert assessments

The role of domain experts in DeepER-MedQA is central to the framework. The benchmark is built upon questions derived from real-world biomedical investigations, capturing the types of reasoning required for mechanistic research, translational studies, and clinical decision-making. Expert participation not only guided question design but also shaped evaluation criteria and curation of reference answers.

By incorporating multidimensional assessments of analytical coherence, reference relevance, and novelty of insight, DeepER-MedQA evaluates properties of scientific reasoning that are poorly captured by automated metrics alone. Integrating expert oversight with structured LLM-based verification enables scalable validation while reducing the propagation of hallucinated or weakly supported claims in the reference answers (**Fig. 2**). Moreover, the inclusion of detailed expert annotations facilitates the development of automated evaluation models. Our additional analysis shows that these annotations can be used to calibrate an LLM-based judge, with evaluation trends closely aligned with human judgments across systems (Pearson correlation score reaches 0.981).

The clinical case study further illustrates the potential role of evidence-based generation in real-world decision environments (**Tab. 1**). Precision oncology tumor board discussions require rapid synthesis of heterogeneous information, including clinical histories, genomic variants, biomarker profiles, and emerging therapeutic evidence, from rapidly expanding literature. Unlike benchmark tasks, these cases require contextual interpretation of patient-specific data and alignment of literature evidence with clinical decision points. In this setting, the proposed DeepER-Med produced outputs judged consistently with tumor board recommendations in most cases. Although such systems cannot replace multidisciplinary clinical decision-making, they may help clinicians organize complex information and identify relevant evidence during preparatory analysis.

**Limitations and future directions**

Despite these strengths, several limitations remain. Expert evaluation revealed that DeepER-Med occasionally produced outputs with incomplete reasoning or fragmented evidence integration (**Append Table 2**). In some cases, the system assembled multiple relevant studies without fully establishing biological relationships among them, leading to responses that resembled parallel summaries rather than integrated explanations. Other failures reflected misalignment with question constraints or insufficient attention to temporal relevance, particularly when highly cited older studies overshadowed more recent findings. These observations highlight the continuing challenge of transforming retrieved biomedical literature into coherent and context-aware reasoning, particularly in domains where clinical interpretations evolve rapidly and older highly cited studies may overshadow newer evidence.

Additional limitations arise from the scope of the benchmark and evaluation framework. Due to the high cost of expert curation, DeepER-MedQA contains 100 deep research questions spanning diverse biomedical domains, and therefore cannot fully capture the breadth and heterogeneity of real clinical and research problems. However, DeepER-MedQA made it possible to evaluate deep research systems towards real-world performance in biomedical discovery. Manual evaluation also relies on expert judgment, which may introduce domain-specific biases. Moreover, the performance of DeepER-Med depends on the availability and quality of biomedical literature and structured knowledge resources, which vary across fields. Finally, like most existing deep research systems, DeepER-Med requires substantial time to process complex queries, with efficiency comparable to systems such as OpenAI Deep Research.

Future work will expand DeepER-MedQA to larger and more clinically oriented datasets, explore prospective evaluations in real research settings, strengthen modeling of biological associations among multiple pieces of evidence, investigate tighter integration with structured biomedical and clinical data sources, and optimize the parallel process to speed up the answer generation process. More broadly, the results suggest that progress in AI-assisted biomedical discovery may depend less on scaling model architectures alone and more on designing systems that explicitly represent research intent, evidence selection criteria, and interpretive reasoning. Embedding these principles into AI workflows may help ensure that advances in automated reasoning strengthen rather than obscure the evidentiary foundations of biomedical science.

## Acknowledgements

We would like to thank NIH colleagues for their assistance and discussion of this work: Qiao Jin, Qingqing Zhu, Maya Geer, Yao Ge, Yin Fang, and Ran Gu. This research was supported in part by the Intramural Research Program of the National Institutes of Health (NIH). The contributions of the NIH authors are considered Works of the United States Government. The findings and conclusions presented in this paper are those of the authors and do not necessarily reflect the views of the NIH or the U.S. Department of Health and Human Services. Jordina Rincon-Torroella was supported by the National Institute of Neurological Disorders and Stroke of the National Institutes of Health under Award Number 1K08NS140637. The Hunterian Neurosurgical Laboratory would like to acknowledge the philanthropic support from the Khatib Brain Tumor Center, The Florence D. and Irving J. Sherman MD Foundation Trust, the Venable Foundation and the Jennison Family.

## Author Contributions Statement

**Z.W.** and **Z.L.** conceived this study. **Z.W.** implemented the model construction and evaluation, as well as the result analysis and data curation. **C.W.** developed the knowledge graph retrieval agent and performed the stress test. **J.C.** developed the Streamlit-based platform for expert evaluation. She also developed the API to access ClinicalTrials.gov. **S.T.** and **C.W.** developed the website and agent API for DeepER-Med. **C.-P.D., C.W., M.K., A.F., J.R., H.S., B.T., A.I., M.H., R.L.,** and **T.K.** formulated medical questions and contributed to the answer evaluation and validation. **R.N., N.N.,** and **A.A.** offered clinical questions and contributed to the evaluation. **L.H.** and **N.W.** contributed to collecting responses from OpenAI Deep Research and Open Evidence. **J.C.** and **L.H.** contributed to the human evaluation of open-ended QA task. **Z.W.** and **J.C.** conducted manuscript drafting. **R.L., T.K., L.H.,** and **J.R.** contributed to manuscript revision. **Z.L.** supervised the study. All co-authors reviewed the manuscript and approved the submitted version.

## Competing Interests Statement

All authors declare no competing interests.

## METHODS

### Study Design

In this study, we developed an open-source deep research paradigm that advances evidence-based generation (EBG) for medical research, comprising an evidence-driven agentic framework (DeepER-Med) and an expert-curated benchmark (DeepER-MedQA). DeepER-Med is implemented using GPT-4o (version 20240513) accessed through the Azure OpenAI API, a HIPAA-compliant platform that provides robust data privacy protection[45], together with Gemini-3-Pro accessed via the Google AI platform. GPT-4o queries were executed with a temperature of 0.0 to ensure deterministic and reproducible outputs, whereas Gemini-3-Pro was used with default inference settings to support effective long-context evidence synthesis. In total, DeepER-Med integrates 13 external APIs to retrieve question-relevant evidence from heterogeneous medical resources and to interpret this evidence under transparent and predefined inclusion criteria. DeepER-MedQA was curated through a structured human–AI interaction process, in which GPT-5.2-Pro, accessed via the OpenAI platform, was used to support automated evaluation and reference answer curation.

To assess performance, we manually compared DeepER-Med with three widely used deep research platforms, i.e., OpenAI Deep Research, OpenEvidence, and Google AI Mode, using the DeepER-MedQA dataset. Manual evaluation was conducted across five predefined dimensions: answer accuracy, analytical quality, reference relevance, novelty of insight, and overall comprehension. Additionally, we conducted automatic evaluations of DeepER-Med on five open-access biomedical datasets spanning two task types: question answering and hypothesis verification. Automatic evaluation focused on two complementary objectives: model precision and evidence relevance, which were quantified using absolute accuracy and information-theoretic measures, respectively. The accuracy evaluation for the open-ended QA task was performed by GPT-5.2 (version 20251211), which was accessed through the Azure OpenAI API.

### Statistical analysis

All statistical analyses were performed using Python (version 3.11.0). The framework of DeepER-Med was implemented using requests (v2.32.2), openai (v1.99.7), and google-genai (v1.29.0). Other required dependent packages included google-auth (v2.40.3), tiktoken (0.12.0), pandas (v2.3.2), torch (v2.4.1), and transformers (v4.45.2). Annotations were performed on Streamlit platform (v1.53), and evaluation results were stored in the MongoDB database[46] (v8.0.19). All evidence relevance analyses were performed based on E-utils API (updated on Nov. 17, 2022), ClinicalTrial.gov API (v3.5.0), MedCPT (v20231025), numpy (v1.24.0), seaborn (v0.12.2), matplotlib (v3.7.2) and scipy (v1.10.1). For baseline comparisons, we tested OpenAI Deep Research (access in Dec. 2025), OpenEvidence (access in Dec. 2025), and Google Deep Search (access in Dec. 2025) on DeepER-MedQA using their online platform.

### Overview of evidence-based generation in deep medical research

To conduct trustworthy and transparent deep medical research, we present evidence-based generation paradigm that starts with developing an evidence-grounded agentic AI framework (DeepER-Med).and followed by constructing an expert-level benchmark (DeepER-MedQA).

**Framework of DeepER-Med**

DeepER-Med, an agentic AI system, structures evidence-based medical research as an explicit and inspectable workflow comprising three stages: research intent investigation, evidence retrieval and interpretation, and structured knowledge synthesis. Operationally, the workflow is implemented through three modules, i.e., research planning, agentic collaboration, and evidence synthesis, to support transparent evidence retrieval and integration (**Fig. 1a,b**).

(1) Research planning.

In the research planning module, an input question ($Q$) is decomposed into a set of $k$ atomic and successive sub-questions $\{Q_i\}_{i=1}^{k}$ to explore research intent from multiple perspectives. Sub-questions are generated by GPT-4o, which is selected due to its fast response time and strong contextual understanding[47]. The decomposition typically begins with clarification of core concepts and progressively advances toward deeper associations among concepts, such that the complexity increases from $Q_1$ to $Q_k$, reflecting a realistic progression of scientific inquiry.

After sub-question decomposition, we instruct GPT-4o to assign the most appropriate agents to each $Q_i$ for subsequent evidence retrieval and interpretation within the agentic collaboration module. The agent assignment is guided by a document describing the functional scope of each agent, ensuring that selected agents align with the intent of each sub-question.

(2) Agentic collaboration.

In the agentic collaboration module, DeepER-Med selects appropriate application programming interfaces (APIs) for each sub-question ($Q_i$) based on the documented capabilities of available biomedical tools. These selected APIs are used to retrieve and interpret a relevant evidence set ($E_i$) for the corresponding $Q_i$. To preserve traceability and transparency, we implement a three-layer hierarchical agentic network:

- At the worker layer, DeepER-Med configures 13 APIs to access domain-specific resources, including a medical knowledge graph (i.e., PrimeKG[48]), literature repositories (Wikipedia and PubMed), clinical/drug records, and LLM-based tools. Each API performs a specific function (e.g., entity normalization, abstract retrieval, or citation collection).
    - *get_normalized_entity*: align free-text concepts with the entities in the knowledge graph by using the hybrid of exact match and similarity match.
    - *get_tail_entity_by_relation*: retrieve the associated tail entities from knowledge graph based on the given head entity and the specific relation type.
    - *get_tail_entity_by_type*: retrieve the associated tail entities from knowledge graph based on the given head entity and the required tail entity type.
    - *get_relation_type*: retrieve the most possible relation type of the given entity pair.

- get_shortest_paths: retrieve the shortest paths from the knowledge graph between the given entity pair.
- *get_shortest_path_by_entity_type*: retrieve the shortest paths from the knowledge graph between the given entity pair based on the required entity type.
- *get_pubmed_abstracts*: query PubMed to search abstracts of articles/literature.
- *get_pubmed_full_text*: query PubMed Central to search full text of articles/literature.
- *get_article_citation_details*: search citation information from iCite[49] using PubMed Identifier (PMID).
- *get_wikipedia_introduction*: query Wikipedia to get definitions of biological concepts.
- *get_pubmed_citation_style*: search the standard citation format of a PMID.
- *get_clinical_trials*: query ClinicalTrials.gov to search clinical reports.
- *gemini_response*: generate response for the sub-question using the Gemini model.
- At the manager layer, functionally related APIs are unified into task-oriented communities that execute higher-level operations such as query expansion, evidence retrieval, and LLM responses. Community topology is selected based on task requirements. For example, the query expansion community adopts a tree-like structure because concepts extracted from $Q_i$ may not project directly to knowledge graph entities and therefore require iterative normalization and refinement before graph querying. In contrast, the evidence retrieval community is organized as a network structure to support multi-step integration of literature search, citation tracking, and application of evidence inclusion criteria.
    - *Query Expansion* contains six APIs: get_normalized_entity, get_tail_entity_by_type, get_relation_type, get_tail_entity_by_relation, get_shortest_path_by_entity_type, and get_shortest_paths.
    - *Evidence Retrieval* consists of six APIs: get_clinical_trails, get_pubmed_abstracts, get_pubmed_full_text, get_article_citation_details, get_pubmed_citation_style, and get_wikipedia_introduction
    - *LLM Response* is implemented by the gemini_response API.
- At the director layer, DeepER-Med applies explicit global logical rules to coordinate interactions across three communities and to integrate heterogeneous evidence in a transparent manner. Specifically, for each sub-question $Q_i$, one of the following two processing strategies is selected: If the sub-question requires relational reasoning, the system performs query expansion using the knowledge graph and subsequently retrieves supporting evidence for the expanded queries. Otherwise, it directly parses the sub-question into a set of queries and retrieves evidence without prior expansion. In both cases, the workflow ends with evidence screening and appraisal stage that prioritizes methodological quality, contextual relevance, and evidentiary strength based on the user-defined preference for evidence inclusion. The LLM response component (Gemini-3-pro in our implementation) is activated only when no high-quality evidence is available. Its generated content is retained only if it is subsequently supported by evidence retrieved from source databases. In this design, Gemini-3-pro complements GPT-based components by increasing diversity in evidence exploration and interpretation.

(3) Evidence Synthesis.

In the evidence synthesis module, each $(Q_i, E_i)$ pair is re-interpreted using GPT-4o. During this step, the evidence set is refined to prioritize items most directly supporting the core intent of $Q_i$, while preserving traceability to original sources. Background information embedded in the query, together with user-provided research objectives, is treated as an explicit constraint that narrows the scope of candidate evidence produced by the agentic collaborative network, ensuring that synthesis remains aligned with the researcher's focus. Subsequently, Gemini-3-pro is selected for the evidence synthesis and answer generation owing to its strong long-context reasoning capabilities and empirically observed reliability in generating credible references in medical research[50]. In this way, DeepER-Med produces a concise final answer accompanied by a detailed analytical report, with both outputs explicitly grounded in the supporting evidence. All cited sources are listed in the reference section in order of appearance, enabling efficient review, verification, and downstream analysis by researchers.

**Curation of DeepER-MedQA**

To construct an expert-level benchmark reflecting multidisciplinary frontline medical research, we recruited 11 domain experts from the National Institutes of Health (NIH), Johns Hopkins University (JHU), and the University of Illinois Urbana–Champaign (UIUC). The expert panel comprised researchers with expertise spanning translational medicine, preclinical experimentation, basic medical research, and applied biomedical sciences. All experts participated in the full curation process, including question formulation, answer evaluation, and standard curation.

(1) Question formulation.

Based on the research focus and active projects of participating experts, we formulated 100 research questions spanning 11 medical topics (**Fig. 2a, left**). These questions were formulated either as open-ended inquiries guided by interrogative prompts or as heuristic research questions paired with explicit contextual background (**Appendix Tab. 1**) and were categorized into three medical studies according to their primary scientific intent: basic medicine, clinical medicine, and translational medicine (**Fig. 2a, right**). The categorization criteria were defined as follows:

- Basic medicine: questions focused on disease-related mechanistic understanding that underpins medical research but is not directly situated in clinical practice. Examples include "*What controls intestinal Treg IL-10 production?*" and "*Does immune-mediated selection differ by patient sex and age?*".
- Clinical medicine: questions centered on clinical manifestations, diagnosis, or treatment decisions in patient populations. Examples include "*What is the role of diet in age-related macular degeneration progression?*" and "*What are the earliest imaging features of macular telangiectasia type 2?*".

- Translational medicine: questions addressing bench-to-bedside research such as liquid biopsy technologies, biomarkers, and therapeutic targets. Examples include "*How does cfDNA differ from ctDNA?*" and "*How does liquid biopsy guide treatment strategies?*".

(2) Answer generation.

After compiling the set of medical questions, we respectively queried DeepER-Med, OpenAI Deep Research, OpenEvidence, and Google AI Mode (Deep Search) to generate candidate answers. To minimize bias during subsequent expert evaluation, we normalized outputs from all sources into a common template aligned with the DeepER-Med's format that includes (i) a concise final answer summarizing the key findings, accompanied by supporting PMIDs, (ii) a detailed analytical report in which each substantive claim is explicitly supported by at least one cited source, and (iii) a list of supporting references. Furthermore, we develop a Streamlit-based[51] assessment workflow that masks system identities and assigns responses for review to the author who contributed the corresponding questions. This blinded annotation strategy is designed to support fair, domain-informed, and unbiased assessment of evidence quality and reasoning across systems.

(3) Answer evaluation.

After generating candidate responses for each question, the expert who originally contributed that question conducted a blind evaluation of all answers, with the source model fully masked. Each response was scored across five predefined dimensions—answer accuracy, analytical quality, reference relevance, novelty of insight, and overall comprehension (**Fig. 2c, 2d**).

- Answer accuracy, analytical quality, and reference relevance were assessed using three labels. **High** indicates that the response aligned with the question intent and provided correct conclusions, or indicates logically coherent analysis, or indicates highly relevant references; **Moderate** indicates alignment with question intent but incomplete conclusions, or indicates the plain analysis, or indicates broadly relevant references; **Low** indicates misalignment with question intents and incorrect conclusions, or indicates superficial analysis, or indicates the inclusion of irrelevant references.
- Novelty of insight was annotated using **Yes**, **Maybe**, or **No**, reflecting whether the response introduced new perspectives or highlighted nuanced knowledge that might otherwise be overlooked.
- Comprehension was evaluated on a five-point ordinal scale assessing the breadth of relevant knowledge covered in the response, where **a score of 1** indicates minimal coverage and **a score of 5** indicates near-complete coverage of relevant concepts; **a score of 3** represents an average level of comprehension.

After reviewing the four system outputs for a given question, the contributing author selects one (or more) high-quality response as a candidate reference answer (**Fig. 1c**). For each selected response, the associated references are then examined individually to verify their traceability and to detect fabricated or unverifiable citations (i.e., hallucinations). Candidate answers supported by fully traceable evidence proceed directly to the standard curation step. In cases where hallucinated or unverifiable references are detected, the corresponding response is excluded and replaced with an alternative response exhibiting high accuracy and reliable citation. When no single response satisfies the question, we use GPT-5.2-pro[52] with predefined prompting instructions to synthesize multiple high-quality alternatives into a consolidated reference answer for downstream human curation.

(4) Reference answer curation.

In the reference answer curation stage, domain experts and LLM judgment (GPT-5.2-pro) jointly validate the correctness and evidence grounding of each reference answer against the raw research question (**Fig. 1c**). Specifically, domain experts assess factual accuracy, logical coherence, and evidence alignment, while GPT-5.2-pro provides sentence-level judgements of consistency and evidentiary support using predefined criteria. If both expert review and LLM judgment confirm that the reference answer is correct and sufficiently supported by traceable evidence, it is accepted and incorporated into the curated benchmark. If either assessment identifies deficiencies (e.g., insufficient evidence support, logical inconsistencies, or residual hallucinations in the reference list), the answer is returned for revision. In the revision step, the answer is iteratively refined based on expert feedback and re-evaluated through the same human–AI validation loop until consensus is reached. This iterative curation process ensures that all benchmark answers meet stringent standards of accuracy, transparency, and evidence traceability. To mitigate evaluator bias, all assessments followed criteria-based guidelines that prioritize evidence traceability, factual accuracy, and logical coherence over stylistic or subjective considerations. This process resulted in a curated set of reference answers (**Fig.2b**) that met predefined standards of accuracy, transparency, and evidence traceability.

## Automatic evaluation for DeepER-Med

Besides manual evaluation on DeepER-MedQA, we also assessed DeepER-Med at distinct stages of evidence-based medical research on five open-access medical datasets across a QA task and a hypothesis verification task. We set two metrics including accuracy and information entropy to measure the performance of DeepER-Med in terms of model precision and evidence relevance.

### Accuracy for Model precision

We employed two QA datasets for model precision evaluation: BioMaze and PubMedQA. For BioMaze tasking with binary QA and PubMedQA, we directly compare DeepER-Med's outputs ($pred$) with ground-truth labels ($ref$). The accuracy calculation is as follows:

$$accuracy = correct/total \tag{1}$$

where $correct$ is 1 if $pred$ is same as $ref$; else $correct$ is 0. The $total$ denotes the total number of questions in the dataset.

For BioMaze tasking with open-ended QA, we leverage GPT-5.2 along with predefined prompting instructions as an independent evaluator to compare DeepER-Med's outputs and the ground-truth answers. Then, we calculated the proportion of "consistent" decisions in all questions as the accuracy.

**Evidence relevance**

We introduced information entropy to measure the relevance of evidence retrieved by DeepER-Med and curated by experts. We first collected title ($t$) and abstract ($a$) of retrieved or curated literature, and then, we used MedCPT to encode them as 768-dimensional tensors for calculation of information entropy. This process can be formulated as follows:

$$\mathbb{V}_E = MedCPT(E_{t\&a}, \Theta) \tag{2}$$

where $E_{t\&a}$ denotes the evidence with title and abstract. $\Theta$ is the parameter set of MedCPT. $\mathbb{V}_E$ is a 768-dimensional tensor. For each $\{\mathbb{V}_E\}_{i=1}^{768}$, we choose $K = 100$ bins $(b_1, \ldots, b_K)$ for the computation of histogram counts, i.e., $c_k = \#\{i: x_i \in b_k\}$. Then, we convert the histogram counts to a discrete probability vector as shown in Eq. (3).

$$p_k = \frac{c_k}{\sum_{j=1}^{K} c_j} = \frac{c_k}{768} \tag{3}$$

Subsequently, the $p_k$ is stabilized with an adjustment parameter, i.e., $\tilde{p}_k = p_k + \varepsilon, \varepsilon = 10^{-12}$. Finally, the Shannon information entropy is calculated by the Eq. (4).

$$H_2(\mathbb{V}_E) = \frac{-\sum_{k=1}^{K} \hat{p}_k \log(\hat{p}_k)}{\log 2} \tag{4}$$

$H_2(\mathbb{V}_E)$ indicates the information entropy in bits (log base 2), where the $\hat{p}_k = \frac{\tilde{p}_k}{\sum_{j=1}^{K} \tilde{p}_j}$ denotes the empirical probability of the $k$-th histogram bin, obtained by normalizing the bin counts.

Furthermore, to measure the distance between distributions, formally, given two sets of evidence: one retrieved by DeepER-Med (or baselines) and the other one curated by experts, we constructed empirical probability distributions from their corresponding MedCPT embedding representations. Let $P$ and $Q$ denote the resulting distributions, respectively. We quantified the distributional similarity between retrieved and expert-curated evidence using the Jensen–Shannon distance, a symmetric and bounded divergence measure:

$$JSD(P, Q) = \sqrt{\frac{1}{2}KL(P \parallel M) + \frac{1}{2}KL(Q \parallel M)}, \ M = \frac{1}{2}(P + Q) \tag{5}$$

where $M$ is the average distribution of $P$ and $Q$, and $KL(\cdot)$ denotes the Kullback–Leibler divergence. The Jensen–Shannon distance compares each distribution to their shared average, ensuring symmetry and numerical stability, with values bounded in (0, 1). Smaller Jensen–Shannon distance values indicate greater similarity between retrieved evidence and expert curation, with values closer to 0 corresponding to higher alignment, while values closer to 1 reflect greater distributional differences. Additional background on information entropy is provided in **Ref. 26** and **Ref.27**, while Jensen–Shannon distance can be referred at https://medium.com/@vibhorkashyap/understanding-jensen-shannon-distance-a-friendly-guide-for-data-scientists-4cac664c3381.

**Clinician's assessment for case study**

Three clinicians from UIC participated throughout the entire case study process, including clinical case construction, question formulation, and evaluation of DeepER-Med-generated conclusions.

First, the clinicians curated eight clinical cases from tumor board records. Each case included comprehensive clinical information, such as patient history, biomarkers, tumor specimens, genomic variants, biologically relevant features, and immunotherapy markers. All identifiable patient information (e.g., names, dates, and locations) was fully de-identified prior to use. An example case is provided below.

---

**Case 1:** xxxxx

**Diagnosis:** NSCLC

**Clinical History:**

- xxxxx: s/p chemoRT to RUL/hilum/med with SRS to R parietal lesion
- xxxxx: SRS to new brain mets
- xxxxx: developed polymetastatic recurrence, began pembro monotherapy
- xxxxx: with stable systemic disease. New brain MRI showed >50 brain mets, very symptomatic, getting WBRT

**Biomarker of interest**: METex14; PDL1 >80%

**Tumor specimen:** Lung R Hilar

**Testing platform:** Tempus xT

*Genomic variants:*

*Potentially Actionable:* **MET** c.3026_3028+1del Splice region variant Exon 14 deletion –GOF (VAF 28.8%)

*Biologically Relevant*: TP53 p.Y327* Stop gain – LOF (VAF 12.8%)

**Pertinent Negatives**: EGFR, KRAS, BRAF, ALK, ROS1, RET, ERBB2

*Immunotherapy markers:*

**TMB**: 3.2 m/MB (Low – 33rd percentile)

**MSI Status**: Can not be assessed

---

Second, based on the detailed clinical records, the clinicians formulated corresponding clinical questions involving cancer diagnosis, progression, or risk assessment. To ensure compatibility with the model input format, we reorganized the structured and fragmented clinical fields into coherent natural language narratives. These narratives, together with the clinician-formulated questions, were used as inputs to query DeepER-Med, generating corresponding responses.

Finally, the three clinicians jointly reviewed the responses generated by DeepER-Med and conducted an expert evaluation based on two criteria:

- **Consistency with POTB recommendations**: Whether conclusions generated by DeepER-Med align with the recommendations from POTB. This was assessed qualitatively and categorized as either "consistent" or "discordant".
- **Evidence reliability**: Whether the evidence provided by DeepER-Med sufficiently covers well-established and relevant published studies. This assessment was based on consensus among the clinicians and categorized as either "complete" or "partial".

# Extended Table and Figures

**Extended Table 1 (Related to Figure 2b). A complete case of DeepER-MedQA.**

| Category | Clinical Medicine |
| --- | --- |
| Question | What are the earliest imaging features of macular telangiectasia type 2? |
| Reference Answer | The earliest imaging features of macular telangiectasia type 2 (MacTel 2) can be subtle and may include both neurodegenerative/structural changes and early vascular abnormalities. The earliest signs can be detected in asymptomatic fellow eyes of patients with asymmetric disease and include temporal retinal thinning and the presence of scattered, punctuation hyperreflective dots in the outer retinal layers on spectral-domain optical coherence tomography (SD-OCT) (PMID:32804830). Advanced imaging with adaptive optics scanning laser ophthalmoscopy (AO-SLO) reveals unique dark regions in the cone mosaic and a decreased cone density, which are considered early neuronal changes that can occur even in areas without vascular leakage on fluorescein angiography (FA) (PMID:21642620). Other very early morphologic changes include a reduced directional cone reflectance and an asymmetric foveal pit with focal temporal thinning (PMID:26618805). On SD-OCT, inner retinal hypo reflective cavities and clustered hyperreflective foci at the foveola can also be considered early-stage characteristics that may precede more severe outer retinal damage (PMID:33024250). These findings collectively support the modern understanding of MacTel 2 as a primary neurodegenerative condition with secondary vascular involvement. |
| Expert Assessment | **Accuracy**: High<br>**Comprehensive**: 5<br>**Analytical Quality**: High (Good explanation of biological concepts; Insightful analysis of different aspects of the question; Strong evidence supporting the core conclusion; Profound summarization of the entire analysis)<br>**Reference Relevance**: High<br>**Novel Insight**: No<br>**Source**: DeepER-Med |
| Reference List | PMID: 21642620; PMID: 26618805; PMID: 33024250; PMID: 32804830 |
| Analytical Report | Available upon request |

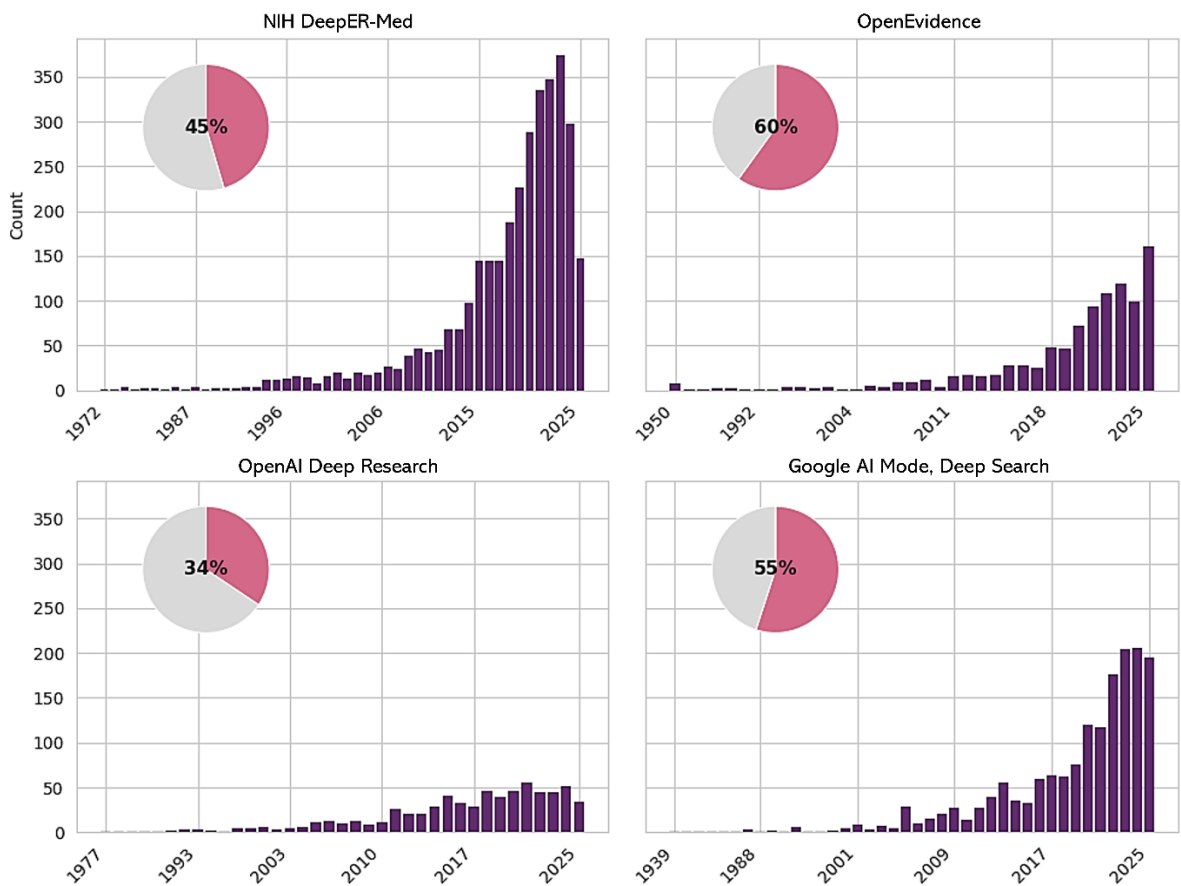

**Extended Figure 1 (Related to Figure 2f). Temporal distribution of literature retrieved by different evidence-aware systems.** Shown is the distribution of publication years for literature retrieved per question by DeepER-Med, OpenAI Deep Research, Open Evidence, and Google Search with AI Models. DeepER-Med retrieves a broader set of literature for each question, with approximately 45% of retrieved studies published within the past five years and the majority spanning from 2015 to 2025. In contrast, OpenAI Deep Research retrieves fewer studies per question, with approximately 34% published within the past five years. Open Evidence similarly retrieves a limited number of studies, although a higher proportion are from recent years. Google Search retrieves a larger volume of recent literature, exceeding DeepER-Med in the proportion of studies from the past five years; however, this set includes a higher incidence of untraceable or unverifiable references.

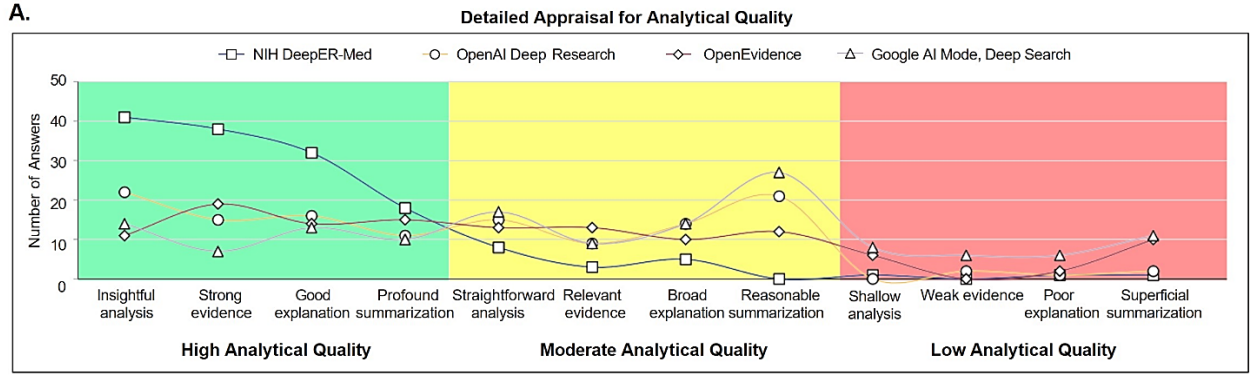

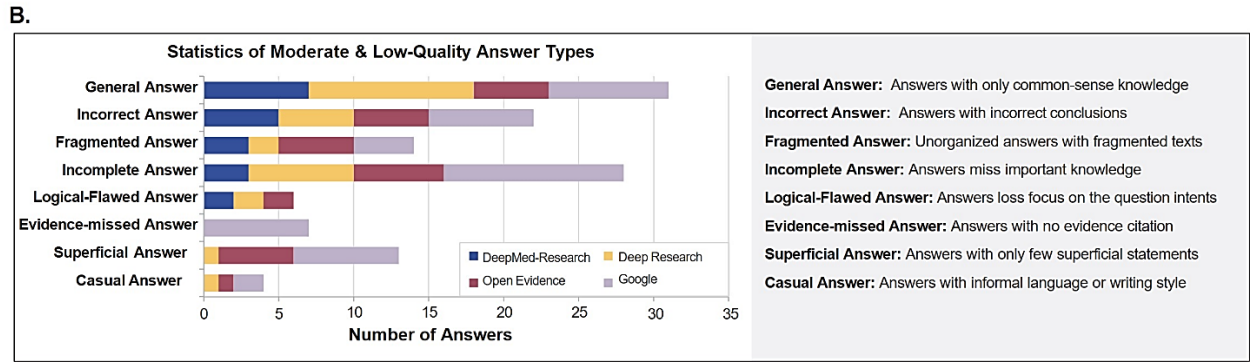

**Extended Figure 2 (Related to Figure 2). Analysis for low-quality answers generated by different deep research systems. A.** The detailed appraisal of experts for the analytical qualities. Each label has four options regarding "analysis", "evidence", "explanation", and "summarization". **B.** The statistics of expert comments on moderate and low-quality answers. The comments are summarized into eight categories (left), while the specific definition of each category is in the right panel.

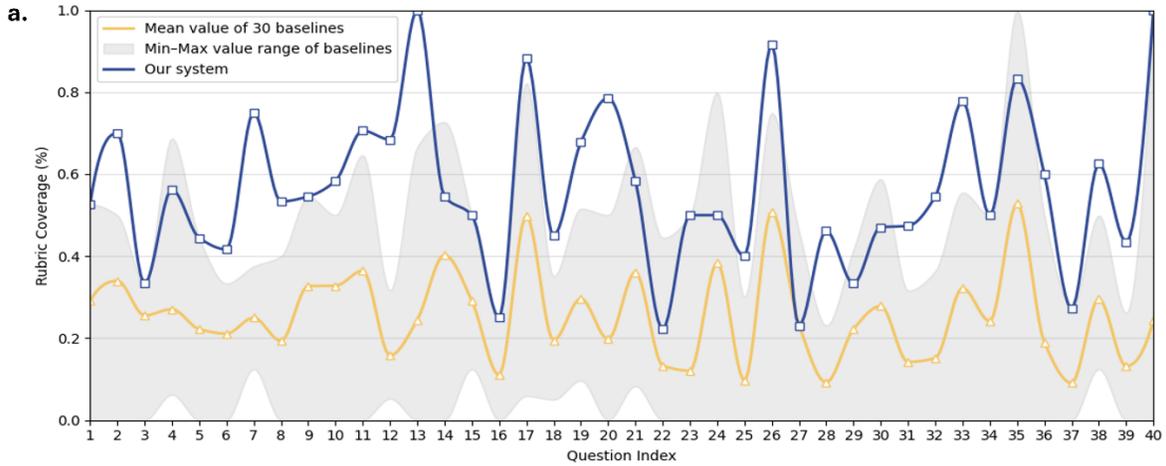

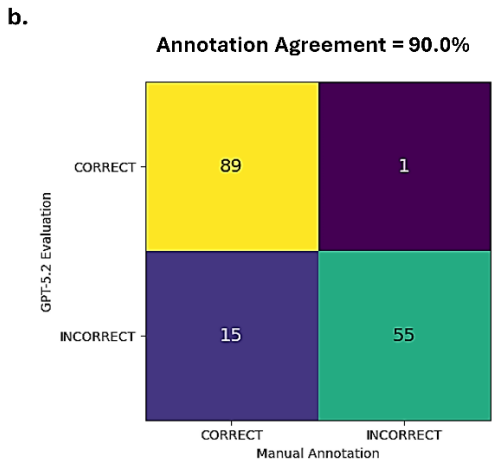
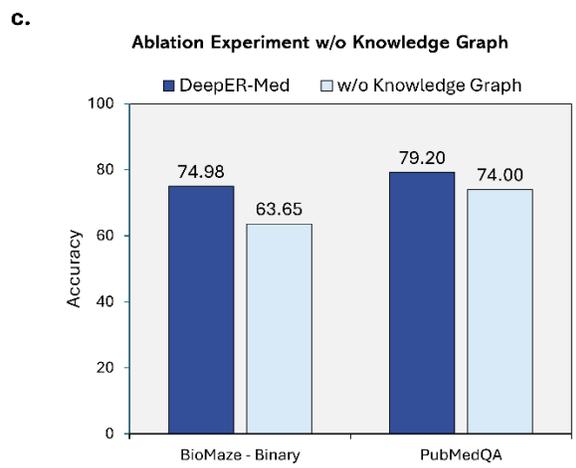

**Extended Figure 3 (Related to Figure 3). Supporting analyses for automatic evaluation of DeepER-Med at different stages of EBG. a,** Comparison of DeepER-Med and 30 baseline models on the MedAESQA dataset, showing that DeepER-Med incorporates a greater number of relevant knowledge elements ("knowledge nuggets") across all 40 questions. **b,** Agreement between automated evaluation (GPT-5.2) and human annotations on the BioMaze open-ended QA dataset, showing 90% concordance and supporting the reliability of GPT-5.2 as an auxiliary evaluator. **c,** Ablation analysis assessing the contribution of query expansion guided by structured knowledge graph. Removal of the knowledge graph component results in performance decreases of approximately 11.3% and 5.2% on two QA datasets, respectively, indicating its importance for model precision.

# Appendix

**Append Table 1. Comparing DeepER-Med with existing AI agents.** Five dimensions spanning evidence presence, transparency in evidence inclusion, medical-oriented research, research-level agents, and expert-aligned evaluation are included.

|  | Evidence-grounded System | Medical-oriented | Deep Research Paradigm | Traceable Evidence Inclusion | Expert-aligned Benchmarking |
|---|---|---|---|---|---|
| Google AI mode (Deep Search) | ✓ |  |  |  |  |
| OpenAI Deep Research | ✓ |  | ✓ |  |  |
| OpenEvidence | ✓ | ✓ | ✓ |  |  |
| Biomni | ✓ | ✓ |  | ✓ |  |
| DeepEvidence | ✓ | ✓ |  | ✓ |  |
| Alvessa | ✓ |  |  | ✓ |  |
| NIH DeepER-Med | ✓ | ✓ | ✓ | ✓ | ✓ |

**Append Table 2**: Examples of error analysis.

| Error Type | Question | Answer generated by DeepER-Med |
|---|---|---|
| Assemble of fragmented pieces | What are the signaling that regulate T cell activity through IKZF1 and IKZF3 in the tumor microenvironment? | In the tumor microenvironment, T cell activity is regulated through the transcription factors IKZF1 (Ikaros) and IKZF3 (Aiolos) primarily via the Cereblon (CRBN) E3 ligase signaling pathway, which is targeted by immunomodulatory drugs (IMiDs). These drugs induce the degradation of IKZF1 and IKZF3, which act as transcriptional repressors. This degradation enhances anti-tumor T cell activity through several mechanisms: 1) de-repression of the Interleukin-2 (IL-2) gene, leading to increased IL-2 production and T cell co-stimulation; 2) upregulation of Granzyme-B (GZM-B), which boosts T cell cytotoxicity; and 3) upregulation of MHC class I and II expression on tumor cells, increasing their visibility to T cells **[Low logical coherence]**. Additionally, IMiDs can directly activate the T cell receptor (TCR) signaling kinase Zap-70, independent of CRBN, to further promote cytotoxicity. Within regulatory T cells (Tregs), the degradation of IKZF1 and IKZF3 disrupts their interaction with the master regulator Foxp3, impairing Treg suppressive function and promoting anti-tumor immunity. Therapeutically, inhibiting IKZF3 function, for instance through knockout in CAR T cells, significantly enhances T cell activation, proliferation, and anti-tumor efficacy against solid tumors. |
| Misalignment with question intent | What determines the difference in the melanoma incidence among **human races**? | The difference in melanoma incidence among human races is primarily determined by the degree of skin pigmentation, which dictates the level of natural protection against ultraviolet (UV) radiation. Individuals with darker skin have higher levels of eumelanin, a pigment that effectively absorbs and scatters UV radiation, thereby reducing UV-induced DNA damage in skin cells and resulting in a significantly lower incidence of melanoma compared to fair-skinned individuals. While UV exposure is the predominant environmental risk factor for melanoma, its role is most pronounced in fair-skinned populations; the association is less certain and often not supported by evidence for people with skin of color. Genetic factors also play a crucial role, with specific genes like MC1R influencing pigmentation and DNA repair capacity, and other susceptibility genes like CDKN2A contributing to risk. However, the overall difference in melanoma burden is compounded by a complex interplay of non-biological factors linked to race as a social construct. These include disparities in socioeconomic status, healthcare access and quality, cultural practices related to sun exposure, and the effectiveness of public health campaigns, which are often not tailored to communities of color. Consequently, while people of color have a lower incidence of melanoma, they frequently experience delayed diagnosis, presentation at more advanced stages (e.g., acral lentiginous melanoma in non-sun-exposed areas), and worse survival outcomes, highlighting a critical disparity where biological advantage is negated by systemic and social factors **[Irrelevant]**. |
| Inclusion of outdated conclusions | What protein kinases are most important in the normal regulation of sodium excretion and what are the manifestations of mutation of these kinases? | The primary protein kinases that regulate sodium excretion are the With-No-Lysine (WNK) kinases (WNK1, WNK4), Serum- and Glucocorticoid-induced Kinase 1 (SGK1), and G protein-coupled Receptor Kinase type 4 (GRK4). WNK kinases and SGK1 form a critical signaling cascade in the distal nephron that controls the activity of the Na-Cl cotransporter (NCC) and the epithelial Na+ channel (ENaC), two key players in sodium reabsorption. This cascade is modulated by hormones like aldosterone and angiotensin II, intracellular chloride concentration, and protein degradation machinery involving the CUL3-KLHL3 ubiquitin ligase complex. GRK4 regulates sodium excretion through the dopaminergic system by phosphorylating dopamine receptors, which impairs their ability to promote natriuresis **[Outdated knowledge]**. Mutations in these kinases disrupt this delicate balance. Gain-of-function mutations in *WNK1* and *WNK4*, or mutations in the regulatory proteins *CUL3* and *KLHL3* that lead to increased WNK kinase levels, cause Familial Hyperkalemic Hypertension (FHHt). This inherited disease is characterized by excessive sodium reabsorption, leading to salt-sensitive hypertension, hyperkalemia, and metabolic acidosis. Specific variants in the *GRK4* gene (e.g., R65L, A142V, A486V) are also associated with essential hypertension and salt sensitivity by desensitizing dopamine receptors and reducing sodium excretion **[Outdated knowledge]**. |